# A Recipe for Social Media Analysis

Shahid Alam
Department of Computer Engineering
Adana Alparsalan Turkes Science and Technology University
Adana, Turkey
salam@atu.edu.tr

Juvariya Khan
Department of Management Sciences
Virtual University of Pakistan
Lahore, Pakistan
juvariya.mak93@gmail.com

*Abstract* — The Ubiquitous nature of smartphones has significantly increased the use of social media platforms, such as Facebook, Twitter, TikTok, and LinkedIn, etc., among the public, government, and businesses. Facebook generated ~70 billion USD in 2019 in advertisement revenues alone, a ~27% increase from the previous year. Social media has also played a strong role in outbreaks of social protests responsible for political changes in different countries. As we can see from the above examples, social media plays a big role in business intelligence and international politics. In this paper, we present and discuss a high-level functional intelligence model (recipe) of Social Media Analysis (SMA). This model synthesizes the input data and uses operational intelligence to provide actionable recommendations. In addition, it also matches the synthesized function of the experiences and learning gained from the environment. The SMA model presented is independent of the application domain, and can be applied to different domains, such as Education, Healthcare and Government, etc. Finally, we also present some of the challenges faced by SMA and how the SMA model presented in this paper solves them.

## I. Introduction

Merriam-Webster dictionary defines social media [1] as, *a form of electronic communication, such as websites and applications used for social networking and microblogging, where users create online communities and groups to share information, ideas, personal messages, and other contents.* Some examples of these platforms are Twitter, Facebook, YouTube, WhatsApp, WeChat, TikTok, LinkedIn, Wikipedia, blogs, and news. The ubiquitous nature of smartphones has significantly increased the use of these social media platforms among the general public. In addition to the general public these platforms are also, used by governments: to interact with citizens, analyze/monitor public opinion and activities, etc. [2]; and use by businesses: for e-commerce, marketing research, communication, sales promotions and discounts, and employee learning, etc.

Modern businesses consider social media as a promising platform to effectively communicate with the targeted customers for conducting marketing and other promotional activities. As an example, Facebook (some of its popular social media subsidiaries also include WhatsApp and Instagram) generated ~70 billion USD in 2019 in advertisement revenues alone, a ~27% increase from the previous year [3]. This indicates, that social media platforms have become a significant portion of digital marketing and will play a major role in the coming years as well.

The way people use and perceive social media defines their powers and practicalities. The use of social media can empower ordinary citizens to conceive ways to sustain the crisis. Coronavirus has restricted face to face social interactions, but due to the innovative use of social media (Facebook and Twitter), farmers in India can connect directly with their end customers, and are finding a way out of this crisis [4]. Social media has also played a strong role in outbreaks of social protests responsible for political changes in different countries [5].

As we can see from the above examples, social media plays a big role in business intelligence and international politics. Although social media contains text, images, and videos, for this paper we restrict the contents to only textual data and discuss the recipes (methods and procedures) only for analyzing the textual data.

Social media analysis (SMA) is a set of techniques and tools to accumulate, aggregate, and systematize social media data to discover and provide significant patterns. The content found on social media platforms typically yields semi-structured or unstructured text. This type of text lacks metadata and is difficult to analyze directly. It typically contains emails, instant messages, and documents, that are user-generated. Since the bulk of text on social media is semi-structured or unstructured, we need to first convert this to a structured data, to manage and analyze and to make sense out of this text. To make decisions based on this data and its analysis (data-driven decisions) we want to present the analysis in a form that makes it easy to interpret. Techniques from different technical areas are used for this purpose, such as natural language processing (NLP), machine learning (ML), information retrieval (IR), statistics and data visualization (DV).

## II. Ingredients

Before giving the recipe for SMA, we first discuss the core ingredients (major technical areas), their terminologies and list some of their techniques to better understand the recipe. Here we just give a brief introduction to these areas. For a detailed discussion, review, and explanation, the reader is referred to the respective references for each of these technical areas given in this paper.

### A. Natural language Processing

NLP [6], [7] is a branch of artificial intelligence and computational linguistics that deals with the processing of a natural language, such as English, Spanish, French and Chinese, etc. NLP is used to extract important information from natural language input and/or produce natural language output. Different levels/types of language processing in NLP are: phonology – interpreting speech sounds; morphology – words are composed of morphemes, the smallest units of meaning, such as preregistration is composed of the prefix pre, the root registration and the suffixation; lexical – meaning of words; syntactic – uncover the grammatical structure; semantic – meaning of sentences; discourse, meaning of text longer than a sentence; pragmatic – uncover extra meaning, e.g., the context in which the text is used. Different sub-techniques of NLP are used for this purpose. Some of them are, lemmatization, part-of-speech tagging, parsing, morphological and word segmentation, stemming, and lexical semantics. Some of the uses of NLP are, named entity recognition, semantic analysis, opinion mining, topic segmentation, automatic summarization, and text-to-speech.

There is a strong and natural relation between NLP and SMA. Therefore, techniques from NLP form the most important ingredients of SMA.

*B. Machine Learning*

ML [8], [9] is a branch of artificial intelligence that automatically makes predictions based on a mathematical model build from sample data also called training data. Different approaches used for learning from the training data are supervised learning – the learning is performed on labeled sample data; some of the algorithms used are, support vector machines, linear regression, Naive Bayes and decision trees; unsupervised learning – the learning is performed on unlabeled sample data; some of the algorithms used are, clustering and anomaly detection.

The success of a learning algorithm depends on the data used. Therefore, ML techniques are data-driven and combine basic concepts from computer science, statistics, probability ad optimization. Some of the problems solved by ML are text or document classification, spam detection, network intrusion, fraud detection, and optical character recognition.

Some of the terminologies commonly used in ML are:

o Dataset – Items or instances of data that will be used for learning and evaluation.

o Features – The set of attributes/properties associated with each sample in the dataset.

o Labels – Classes or categories assigned to each sample in the dataset.

o Hyperparameters – Parameters that are specified as inputs to the learning algorithm.

o Training sample – Samples from the dataset used for learning.

o Validation sample – Samples from the dataset used for selecting appropriate hyperparameters. They can be part of the training sample.

o Testing sample – Samples from the dataset used for evaluating the performance of a learning algorithm.

o Hypothesis set – A set of functions (algorithms) mapping features to the set of labels.

The first step in ML is to randomly divide the dataset into training, validation, and testing sets. Next comes the association (selection) of features to samples in the dataset. The selected features are used to train the learning algorithm and tune the hyperparameters. For each value (set) of these parameters, a different hypothesis is selected from the set of hypotheses. In the end, the best performing hypothesis on the validation sample is selected. Finally, the prediction is made on the testing sample using the selected hypothesis. The performance of the algorithm is evaluated using different metrics. Techniques from ML also form the most important ingredients of SMA.

*C. Information Retrieval*

IR [10] is defined as a set of techniques that are used for finding information in the form of data (usually text) of an unstructured or semi structured nature that satisfies an information requirement from within large sets of data (documents). This large set of data is referred to as the collection or a corpus (a body of text). Some of the important IR models are Boolean, Vector Space, and Latent Semantic Indexing.

A Boolean Model is a model in which documents are set of terms and queries are Boolean expressions. A typical Boolean model contains: a set of words, i.e., the indexing terms either present (1) or absent (0); a Boolean expression; Boolean algebra; a prediction, a document is predicted as relevant if it satisfies the query. Some of the properties of a Boolean model are: it only retrieves exact matches; it is a very simple model based on sets and is easy to implement; it does not provide the ranking of the retrieved documents.

A Vector Space Model represents a set of documents as vectors in a common vector space. The queries are considered as vectors in a high dimensional Euclidean space. The similarity of vectors (a document vector and a query vector) is computed using the cosine similarity (cosine of the angle between them). Different techniques are used to rank the retrieved document, such as term frequency, document frequency, and collection frequency. Some of the properties of a vector space model are: it is a simple model based on linear algebra; it allows partial matching; it provides the ranking of the retrieved documents.

A Latent Semantic Indexing Model is an extension of the vector space model and analyzes the relationship between a set of documents and the terms (words) they contain. Latent semantic indexing assumes that words that are close in meaning will occur in semantically similar texts. A term document matrix is used to store these similarity values, which can be queried to retrieve specific documents. A term document matrix is a 2D matrix that lists the frequency that each term occurs in the documents. The terms are assigned weights using TF-IDF, which assigns more weight to the rare terms. To lower the dimensions and noise, a low-rank approximation [10] is applied to the term-document matrix. This new low-dimensional matrix is then queried to retrieve specific documents.

IR models select and(or) rank the document that is retrieved by a query. Therefore, techniques from IR are mostly used for accumulating the data and form an important part of the ingredients of SMA.

*D. Data Visualization*

The ability of humans to remember pictures (visuals) far better than words/texts [11], [12] makes humans more visually inclined. It is easy for them to understand and consume certain information in a visual form than words/text. Therefore in today's data-driven world, DV [13] is very critical to efficiently and aptly communicate with humans. DV deals with representing the data graphically. A graphic mark represents a data value or a set of data values. This mapping helps communicate information clearly and efficiently to the users. We can divide these quantitative mappings and relationships into seven types [14]:

o Time-Series – Instances of one or more measures at equidistant points. A line chart may be used to represent such a relationship.

o Correlation – Comparison between two variables to see if they tend to move in the same or opposite direction. A scatter plot may be used to represent such a relationship.

- o Ranking – Data values ordered by size or intensity. A bar chart or heat map may be used to represent such a relationship.
- o Part-to-Whole – Data values representing parts of a whole. A pie or bar chart may be used to represent such a relationship.
- o Deviation – Data values compared against a reference. A bar chart may be used to represent such a relationship.
- o Distribution – Count of a variable in a given interval. A histogram or a boxplot chart may be used to represent such a relationship.
- o Geospatial – Comparison of data values across a map or location. A network chart may be used to represent such a relationship.
- o Nominal Comparison – Comparison of discrete values with no particular order. A bar chart may be used to represent such a relationship.

A typical data visualization process includes: Data Import – extract data from the source; Data Preparation – prepare the data for visualization, e.g., normalizing values and interpolating missing values; Data Manipulation – select the data to visualize, e.g., filtering, joining and grouping; Mapping – map the data to geometric primitives, e.g., points and lines; Rendering – transform the geometric data into visual depiction. Techniques from DV form the final critical ingredients of SMA.

Figure 1 lists some of the techniques used in each of these areas. We only listed the core techniques here. The reader interested in getting more details and explanations about these and other such techniques is referred to the respective references given in this paper for each of the technical areas.

### III. RECIPE (MODEL)

We present in Figure 2 a high-level functional intelligence model (recipe) of SMA that synthesizes the input data and uses operational intelligence to provide actionable recommendations. This model also matches the synthesized function of the experiences/learning gained from the environment. The model/process of SMA is divided into three steps (components).

1. Data Identification — Data analysis is as good as the data we are searching in. Therefore it is very important to identify the proper data sources to evaluate. In this step we are basically answering the questions about whose opinions or ideas/thoughts we are interested in, where the conversations are happening, and do we need to just look in the current or past (when) conversations.

2. Data Analysis — After collecting the data, now we want to answer some questions related to the data, such as what are the opinions of customers about a certain product of a company, or how to rate the evaluations/ratings of public about a politician running for the presidency of a country. As mentioned before social media platforms yield semi-structured or unstructured text data. We first convert this to structured data and then perform analysis making sense out of this data and trying to answer the questions related to the data.

3. Information Interpretation — So now we have performed some analysis on the data. How to interpret this analysis, or present it in a form that helps people make some important decisions. For example when, where and what information to disseminate to increase the exposure of the prospective customers to a company's products.

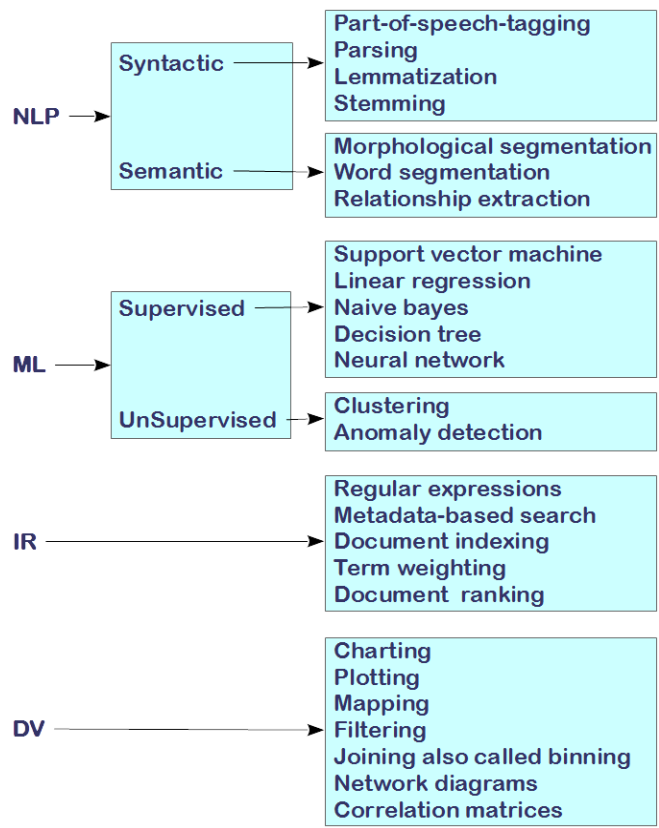

Fig 1. List of some of the core techniques used in each of the technical areas (NLP = Natural language processing, ML = Machine learning, IR = Information retrieval and DV = Data visualization) discussed in section II. Reader interested in getting more details and explanation about these and other such techniques is referred to the respective references given in this paper for each of the technical areas.

Depending on the type of analysis, the SMA model shown in Figure 2 uses one or a combination of techniques from the technical areas discussed in section II. Some of these core techniques are listed in Figure 1. A complete use case of this SMA model is given below.

For example, to increase the sales of a company we would like to analyze and evaluate the customers' and consumers' views about the company's products and services. As the first step in the SMA model presented in this paper, we will identify the data. Depending on the presence of the company on different social media platforms, this step may include retrieving relevant information from Twitter and Facebook, etc. In the second step we will perform data analysis. It may involve performing opinion mining and trend analysis using different (syntactic and semantic) NLP techniques. Different ML techniques will be used to learn from the opinions and trends of the customers and consumers. As a result of this, we will get the most important opinions and trends, which will be used by the company's executives and managers to make important decisions to increase the sales of the company. As the last step, the results will be visualized, using different visualization techniques, to ease the comprehension and better decisions by executives and managers. These decisions may involve changing, updating, and removing some of the company's products.

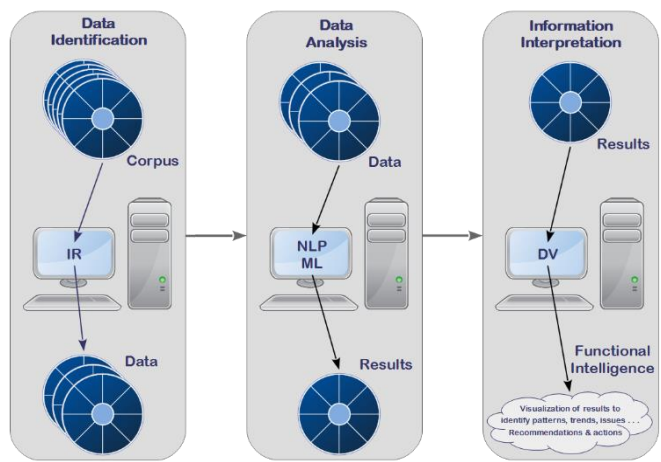

Fig 2. A functional intelligence model of Social Media Analysis divided into three major components, data identification, data analysis and information interpretation. IR = Information Retrieval, NLP = Natural Language Processing, ML = Machine Learning and DV = Data Visualization.

*A. Applications*

In this section, we give some applications of the SMA model, i.e., how this model can be used in some of the different domains. Here, we just describe three of them.

a) Education: For data identification and retrieval, we would like to take observations, interviews, and the analysis of a participant's products such as articles, blogs, and writeups on social sites, opinions shared on, for example, Twitter, Facebook, and Instagram. For instance, if a school administration wants to find the difficulties the teachers and students are facing in the old curriculum, what are their pain points and which topics are causing worry. Also, we can target the audience from specific groups such as students from schools, colleges, or universities. We can analyze this data using different techniques as described in Figure 1. This way, at a much quicker rate we might get the feedback than the traditional process of adapting a lot of evaluations. As a result, we will get opinions for those topics which are causing frustration for students and teachers. These opinions and issues will be used by higher authorities in the administrative department to visualize whether all the questions and considerations have come to a productive point before planning new strategies. Similarly, SMA can also help in determining the topic of interests that can help educators to develop more personalized learning programs.

b) Healthcare: SMA can be applied for enhancing the health and safety of the public and to support important issues regarding healthcare, welfare, and social safety of the society. As an example, for data identification and retrieval, public health organizations can track posts that talk about the health of people and can get a sense of the severity of the symptoms in real-time. In another case, we can track conversations to know how the public is responding to emergent health issues. To get the general public thinking we would use social data while applying SMA. By monitoring the public talks, we can positively respond to a situation for instance, if we are facing an outbreak, so by putting more resources to deal with spreading health problems. As a final step, social data gives a track to collaborate and finalize our findings and to measure perception about the illness. We can analyze this data using different techniques as described in Figure 1 and get a real-time response from the health authorities. SMA can be used by health professionals to connect and engage their audience in real-time. Visualizing this analysis, they can influence and build strong relationships with their audience.

c) Government: SMA has been successfully applied by governments to improve governance and to create a strong relationship with its citizens. For data identification retrieval, governments can build social communities such as a page on Facebook and a handle on Twitter, etc. This way conversations can be monitored to create better relationships between the government and citizens. As an example, a government put forwards a proposal and take a referendum approach where voters are asked to vote. For this purpose government would provide a questionnaire or at least two responses for yes or no just to analyze how intelligible the question is and how people would respond to these referendum questions. Easy to understand, clear, and to the point. The proposal will be analyzed and measured, such as considering a time frame, unit of measure, and what factors should be included. Finally, gathered data would be sorted out in a different way such as plotting or finding correlations or creating tables on an Excel sheet. There are other visualizing tools that are helpful in this context. Lastly, the researcher would interpret the findings whether the collected data answered the required question, or helped in defending objections and how. These questions would be of great help for the higher authorities for decision making, approaching better strategies, and to decide the best course of action.

## IV. CHALLENGES FACING SMA

There are various applications and many advantages of SMA as discussed in this paper, but implementing a successful SMA is full of challenges. In this section, we briefly discuss some of these challenges and their solutions.

*A. Unstructured Data*

Unstructured data [15] is the raw data that is not categorized and is in different formats. It is commonly found in text messages. Making sense out of text (words not numbers) is very challenging. For example the word bad may mean good depending on the context, relationship and other variables. A correct semantic analysis, including word segmentation and relationship extraction, is required to properly understand the meaning of a word in such a sentence. Other techniques such as sentiment analysis, pattern recognition, and cognitive analysis are very helpful in solving such problems. To solve some of the problems with unstructured data the SMA model presented in this paper uses such techniques from NLP and ML technical areas.

*B. Real-time Analytics*

One of the main purposes of SMA is to provide quick (real-time) meaningful insights into the data that percolates into an organization to affect strategic planning. Moreover, some of the data on social media changes instantaneously, especially when an event is unfolding. So to be effective, organizations, governments, and businesses should base their plans on real-time SMA [16], [17]. A high signal to noise ratio in social media data makes it difficult to perform and provide real-time analysis. Some of the solutions to this problem are, to use high-end machines, distributed systems, such as cloud computing, and parallelize the analysis as much as possible. The SMA model presented in this paper can be easily ported to a cloud computing environment, and the techniques used can also be parallelized to improve the speed of computation. Moreover, some of the social media platforms provide APIs, such as Streaming API from Twitter, to retrieve data in real-time.

*C. Big Data*

The data on social media shares some of the same characteristics as of big data. Some of the common challenges of big data are [18]: volume – the space required for storage; velocity – the speed of data creation; variety – taking many different forms; veracity – data integrity and authenticity; The solution to some of these challenges are already provided by big data analysis techniques [19] and can also be applied to data on social media.

*D. Data Quality*

Evaluating the quality of the data on social media is always a concern for SMA. A lot of times the information provided is fake. Also, social media is cluttered with fake and duplicate accounts/profiles. Furthermore, the data may not reliable because of some exaggerations or extra information added (to make it more interesting for the readers) to the data. We can use similar techniques for evaluating the quality (believability, validity, and relevancy, etc.) of data on social media as applied to big data [20].

*E. Visualization*

It is very crucial to visualize the data properly when correct decisions need to be taken quickly and efficiently. For effective disaster recoveries, a clear and concise representation of the data needs to be presented to the decision-makers in emergency management services. Traditional visual analysis methods only deal with structured, low volume, and uniform data. The volume, variety (different formats such as textual and geo-data), and fast-changing rate of social media data make it difficult to visualize. Recently different interactive and multimedia techniques [21]–[23] have been proposed to solve this problem. Social media data is noisy and lacks quality. Therefore, visual analytics methods [24] that help provide information about uncertainty and other quality issues will be in high demand.

V. CONCLUSION

In this paper, we have highlighted the importance of SMA and presented a functional intelligence model to successfully implement SMA in any domain. We have also described its applications in different domains and some of the challenges faced by SMA. Currently, we are working on implementing this model. In the future, we will carry out an empirical study to apply this model in different domains and evaluate its effectiveness and performance using various metrics.